\documentclass[conference]{IEEEtran}
\IEEEoverridecommandlockouts
\usepackage{cite}
\usepackage{amsmath,amssymb,amsfonts}
\usepackage{algorithmic}
\usepackage{graphicx}
\usepackage{textcomp}
\usepackage{xcolor}
\def\BibTeX{{\rm B\kern-.05em{\sc i\kern-.025em b}\kern-.08emT\kern-.1667em\lower.7ex\hbox{E}\kern-.125emX}}
    
\usepackage{times}
\usepackage{xcolor}
\usepackage{soul}
\usepackage[utf8]{inputenc}
\usepackage[small]{caption}

\usepackage{latexsym} 
\usepackage{booktabs}
\usepackage{amsmath}
\usepackage{verbatim}
\usepackage{amssymb}
\usepackage{amsthm}
\usepackage{algorithm}
\usepackage{algorithmic}
\usepackage{graphicx}
\usepackage{subfigure}
\usepackage{natbib}
\usepackage{epstopdf}
\usepackage{lipsum}
\usepackage{color}

\usepackage[utf8]{inputenc} 
\usepackage[T1]{fontenc}    
\usepackage{url}            
\usepackage{booktabs}       
\usepackage{amsfonts}       
\usepackage{nicefrac}       
\usepackage{microtype}      

\usepackage{multicol}

\usepackage{url}
\PassOptionsToPackage{bookmarks=false}{hyperref}

\usepackage{hyperref}       
\usepackage{url}            

\begin{document}
\title{Contextual Bandit with Adaptive Feature Extraction}

 \author{\IEEEauthorblockN{1\textsuperscript{st} Baihan Lin}
\IEEEauthorblockA{
\textit{Columbia University}\\
New York, NY, USA \\
baihan.lin@columbia.edu}
\and
\IEEEauthorblockN{2\textsuperscript{nd} Djallel Bouneffouf, Guillermo A. Cecchi,  Irina Rish}
\IEEEauthorblockA{
\textit{IBM Thomas J. Watson Research Center}\\
Yorktown Heights, NY, USA \\
\{dbouneffouf, gcecchi, rish\}@us.ibm.com}
}
 
 

\maketitle

\begin{abstract} 
 We consider an online decision making setting known as  contextual bandit problem, and   propose an approach for improving contextual bandit performance by using an adaptive feature extraction (representation learning) based on online clustering.   Our approach starts with an off-line pre-training on unlabeled history of contexts (which can be exploited by our approach, but not by the  standard contextual bandit), followed by an online selection and adaptation   of encoders.  Specifically, given an input sample (context), the proposed  approach selects the most appropriate encoding function to extract a feature vector which becomes an input for a  contextual bandit, and updates both the bandit and the encoding function  based on the context and on the feedback (reward).  Our experiments on a variety of datasets, and both in stationary and non-stationary environments of several kinds  demonstrate clear advantages of the proposed adaptive representation learning   over the standard contextual bandit based on "raw" input contexts. \footnote{The data and codes to reproduce all empirical results can be accessed at \href{https://github.com/doerlbh/ABaCoDE}{\underline{https://github.com/doerlbh/ABaCoDE}}.}
\end{abstract}


\section{Introduction}
Sequential decision making is a common problem in many practical applications  where the agent must choose the best action to perform at each iteration in order to maximize the cumulative reward over some period of time.
One of the key challenges is achieving a good trade-off between  the exploration of new actions and the exploitation of known actions. This exploration vs. exploitation trade-off in sequential decision making problems is often formulated as the {\em multi-armed bandit (MAB)} problem: given a set of bandit ``arms'' (actions), each associated with a fixed but unknown reward probability distribution \citep{LR85,UCB}, an agent selects an arm to play at each iteration, and receives a reward, drawn according to the selected arm's distribution, independently from the previous actions.

A particularly useful version of MAB is the {\em contextual multi-armed bandit (CMAB)}, or simply the {\em contextual bandit} problem, where at each iteration,  before choosing an arm, the agent observes an $N$-dimensional {\em context}, or {\em feature vector}.
Over time, the goal is to learn the relationship between the context vectors and the rewards, in order to make better prediction which action to choose given the context \citep{AgrawalG13}, as an important problem setting in reinforcement learning \citep{lin2020unified,lin2020astory,lin2020online,lin2019split,bouneffouf2017bandit}.

For example, the  contextual bandit approach is commonly used in various practical sequential decision problems with side information (context), from clinical trials \citep{villar2015multi}, to speaker recognition systems \citep{lin2020speaker,lin2020voiceid}, to recommender system \citep{MaryGP15}, where the patient's information (medical history, etc.), or a speaker's voice profile, or an online user's purchase profile provide a context for making a better decision about a potential treatment or a speaker identity or an ad to show, and the reward represents the outcome  of the selected action, such as, for example, success or failure of a particular treatment option.

However, in certain real-life applications, before  the online decision-making starts, an agent may   have an access  to a    {\em unlabeled context history} (i.e., contexts without the associated rewards), which can be potentially used as a  prior knowledge to  improve the subsequent online  decision-making.
For instance, in medical decision-making settings, the doctor may have an access  to medical records of different patients, which can be used to gain a better understanding of the  patients population. A different example of unlabeled context history can occur in an online recommender setting, where the system may have some previous information about the users, although the reward feedback (e.g., whether the user clicked on the suggested link or not) might be missing. 

Having an access to unlabeled data makes it possible to  {\em  pre-train some  model of the input (contexts) in an offline mode}, and use it later to improve the online decision making. For example, we can learn an autoencoder  to map the raw inputs into  potentially better representations. Moreover, when the inputs are non-homogeneous, we may want to cluster the unlabeled data and learn separate representations for each cluster. Then, in the online mode, we can decide which representation to use for a given context;  such {\em context-driven representation selection} has a potential to further improve the  subsequent decision-making.  These  representation models (e.g.,   autoencoders) can (and should) continue to be updated online as more contexts become available, especially in {\em nonstationary environments} abundant in practical applications, where  both the context and reward distributions can change in various ways.


Motivated by the above scenarios, we consider here a  contextual bandit setting, called {\em Contextual Bandit with Representation learning and unlabeled History (CBRH)}. In this setting, first of all it is assumed that   (1) there is some {\em  set of unlabeled contexts available for pre-training}, before the online decision-making starts, which allows for an initial clustering and encoder construction;  (2)  the bandit's performance can be improved by {\em learning a good context representation (embedding)}  rather than using the raw input, the (3)  embedding functions  are {\em pre-trained on the unlabeled history} and  {\em adaptively selected (and updated) based on the  context} during the online decision-making. Next, we propose  an algorithm for the above CBRH setting, called {\em Adaptive Bandit with Context-Driven Embeddings (ABaCoDE)},   which implements online, clustering-based encoding selection and learning coupled with Thompson-Sampling approach.
 
We evaluate our approach on several types of nonstationary environments and demonstrated that (1) using embeddings, in general,  considerably improves performance of contextual bandit; and (2) moreover, in several cases, adaptive, context-dependent type of embeddings are much better than just one, ``uniform'' embedding. 

 Overall, the lesson learned is   that the  embedding based approach propose here can be a useful tool for improving the performance of contextual bandit; it is helpful to have an access to some  ``unlabeled'' history of contexts to create a reasonable initial embeddings to start with, and to keep augmenting them with respect to new instances arriving in online mode.


 
 To summarize, our approach has several advantages over 
 the standard  contextual bandit: it can exploit the unlabeled context history to learn useful context representations; it allows for a  flexible, adaptive online selection of context-specific representations,  as well as for continuous learning/adaptation  of such representations.

\section{Related Work}
\label{sec:related}
The multi-armed bandit problem has been extensively studied. Optimal solutions have been provided using a stochastic formulation \citep{LR85,UCB}, a Bayesian formulation \citep{T33,BouneffoufF16,AgrawalG12}, or using an adversarial formulation \citep{AuerC98,AuerCFS02}. However, these approaches do not take into account the context which may affect to the arm's performance.
In LINUCB \citep{Li2010,ChuLRS11} and in Contextual Thompson Sampling style (CTS) algorithms \citep{AgrawalG13,bouneffouf2017context}, the authors assume a linear dependency between the expected reward of an action and its context; the representation space is modeled using a set of linear predictors. This assumption is not used in Neural Bandit \citep{AllesiardoFB14}. 
However, these algorithms assume that the agent can observe the reward at each iteration, which is not the case in many practical applications, including those discussed earlier in this paper.

Authors in \citep{bartok2014partial} studies considering some kind of incomplete feedback called "Partial  Monitoring (PM)", which is a general framework for sequential decision making problems with incomplete feedback that allows the learner, when it is possible, to retrieve the expected value of actions through an analysis of the feedback matrix, both of which are assumed to be known to the learner.

In \citep{gajanecorrupt} authors study a variant of the stochastic multi-armed bandit (MAB) problem in which the rewards are corrupted. In this framework, motivated by privacy preserving in online recommender systems, the goal is to maximize the sum of the (unobserved) rewards, based on the observation of transformation of these rewards through a stochastic corruption process with known parameters.

We can say that our setting is similar to the online semi-supervised learning \citep{lin2020semi,Yver2009,ororbia2015online}, which is a field of machine learning that studies learning from both labeled and unlabeled examples in an on-line setting.   However, in   their setting the true label is received at each iteration, while in our setting a bandit feedback is assumed, i.e., if classification was incorrect, the agent will not know what the correct label was, only that its decision was incorrect.  

\section{Background}

\label{background}

This section  introduces some background concepts  our approach builds upon, such as contextual bandit and Thompson Sampling.\\

\noindent{\bf The contextual bandit problem} \\
Following \citep{langford2008epoch}, this problem is defined as follows.
At each time point (iteration) $t \in \{1,...,T\}$, an agent is presented with a {\em context} ({\em feature vector}) $\textbf{x}_t \in \mathbf{R}^N$
  before choosing an arm $k  \in A = \{ 1,...,K\} $.
We will denote by
  $X=\{X_1,...,X_N\}$  the set of features (variables) defining the context.
Let ${\textbf r_t} = (r^{1}_t,...,$ $r^{K}_t)$ denote  a reward vector, where $r^k_t \in [0,1]$ is a reward at time $t$  associated with the arm $k\in A$.
Herein, we will primarily focus on the Bernoulli bandit with binary reward, i.e. $r^k_t \in \{0,1\}$.
Let $\pi: X \rightarrow A$ denote a policy.  Also, $D_{c,r}$ denotes a joint distribution over  $({\bf x},{\bf r})$.
We will assume that the expected reward is a  linear function of the context, i.e.
$E[r^k_t|\textbf{x}_t] $ $= \mu_k^T \textbf{x}_t$,
where $\mu_k$ is an unknown weight vector (to be learned from the data) associated with the arm $k$. \\

\noindent{\textbf{Contextual Thompson Sampling}}\\
In this setting, we consider the general Thompson Sampling, where the reward $r^{i}_t$ for choosing arm $i$ at time $t$   follows a parametric likelihood function $Pr(r_t|\tilde{\mu}_i)$.  Following \citep{AgrawalG13}, the posterior distribution at time $t + 1$, $Pr(\tilde{\mu}_i|r_t) \propto Pr(r_t|\tilde{\mu}_i) Pr(\tilde{\mu}_i)$ is given by a multivariate Gaussian distribution $\mathcal{N}(\hat{\mu}_i(t+1)$, $v^2 B_i(t + 1)^{-1})$, where
$B_i(t)= I_d + \sum^{t-1}_{\tau=1} x_{\tau} x_{\tau}^\top$, and 
where $d$ is the size of the context vectors ${\bf x}_i$, $v= R \sqrt{\frac{24}{\epsilon} d ln(\frac{1}{\gamma})}$ with $R>0$,  $\epsilon \in ]0,1]$, $\gamma \in ]0,1]$ constants, and $\hat{\mu}_i(t)=B_i(t)^{-1} (\sum^{t-1}_{\tau=1} x_{\tau} r_{\tau})$. At every step $t$, the algorithm generates a $d$-dimensional sample $\tilde{\mu_i}$ from
$\mathcal{N}(\hat{\mu_i}(t)$, $ v^2B_i(t)^{-1})$, for each arm, selects the arm $i$ that maximizes $x_t^\top \tilde{\mu_i}$,and obtains reward $r_t$.

\begin{algorithm}[H]
   \caption{The Contextual Thompson Sampling Algorithm}
\label{alg:CTS}
\begin{algorithmic}[1]
 \STATE {\bfseries }\textbf{Initialize:}    \textbf{for } $i=1,...,k$, $B_i= I_d$, $ \hat{\mu}_i= 0_d, f_i = 0_d$.
 \STATE {\bfseries }\textbf{for } $t = 1, 2, . . . ,T$ \textbf{do}
 \STATE \quad Receive context ${\bf x}_t$
 \STATE {\bfseries }\quad   \textbf{for } $i=1,...,k$,  sample $\tilde{\mu_{i}}$ from the $N(\hat{\mu}_i, v^2 B_i^{-1})$  
 \STATE {\bfseries }\quad Choose arm $i_t= \arg \underset{i\subset I}{max}\ x(t)^\top \tilde{\mu_{i}} $
  \STATE {\bfseries }\quad Receive reward   $r^{i}_t$
 \STATE {\bfseries }\quad
 $B_i= B_{i}+ x_t x_t^{T} $, $f_i = f_i + x_t r^i_t$, $\hat{\mu_i} = B_i^{-1} f_i$
 \STATE {\bfseries }\textbf{end}
   \end{algorithmic}
\end{algorithm}
\section{Problem Formulation}
 \label{sec:statement} Using the notation introduced in the previous section, 
 we now  define our novel bandit setting:  {\em Contextual Bandit with Representation learning and unlabeled History (CBRH)} (outlined in Alg. \ref{alg:CBRH}), based on the following key assumptions.
  
First, we assume that {\em a context ${\bf x}_t \in \mathbf{R}^N $ is mapped into its representation} ${\bf z}
_t \in \mathbf{R}^{N_i}$ using an {\em embedding} function $e_i({\bf x}_t)$, selected from a set $E=\{e_1,...,e_k\}$ of currently available embedding functions.
Second, we assume that {\em the set of embedding functions $E$ can be  modified online}. And third,  an access to {\em a set ${\bf D}$ of unlabeled contexts}, i.e. contexts without the  associated rewards, is assumed.
 This dataset can be used, for example,  for {\em pre-training}  embedding functions $e({\bf x})$.
We then define a set
$\Pi = \cup_{e_i \in E}\{\pi: \mathbf{R}^N \rightarrow A,~ \pi({\bf x}) = \hat{\pi}_i(e_i({\bf x}))\}$ of compound-function policies,
where the function $\hat{\pi}: \mathbf{R}^{N_i} \rightarrow A$ maps $\textbf{z}_t = e_i({\bf x_t})$ to an action in $A$. The objective  is to learn a hypothesis $\pi$ over $T$ iterations maximizing the  cumulative reward.

 

  \begin{algorithm}[H]
	\caption{ The CBRH Problem Setting}
	\label{alg:CBRH}
	\begin{algorithmic}[1]
		\STATE Obtain unlabeled
      set of contexts $\textbf{D}$
      \STATE   Learn a context representation model
        \STATE {\bfseries }\textbf{Repeat}
		\STATE {\bfseries }  \quad $({\bf x}_t,r_t)$ is drawn according to distribution $D_{c,r}$
\STATE \quad Choose encoding $e_i \in E$ 
\STATE \quad
Compute representation ${\bf z}_t=e_i({\bf x}_i$)  
\STATE{\bfseries}  \quad Choose an arm $k_t = \hat{\pi}_I(\mathbf{z}_t)$
				\STATE {\bfseries }  \quad The reward $r^k_t$ is revealed
		\STATE {\bfseries }  \quad  Update  policy $ \pi(\cdot) = \hat{\pi}_i(e_i(\cdot))\}$
			\STATE {\bfseries }  \quad $t=t+1$
		\STATE {\bfseries }  \textbf{Until} t=T
	\end{algorithmic}
\end{algorithm}

\section{Adaptive Bandit with Context-Dependent Embeddings (ABaCoDE)   } 

We now describe an adaptive, context-driven embedding selection  approach to solving the  CBRU problem introduced in the previous section. It has two variants,  based on online- and offline clustering, respectively; the choice is controlled by a Boolean input parameter  $isOnline$ in Algorithm \ref{alg:summary}. Two more inputs include: an unlabeled pre-training dataset 
$\textbf{D}$, as well as the number of embeddings $k$.
The algorithm processes the input contexts sequentially, one by one, but at the end of each  {\em mini-batch} of data it updates the embeddings to reflect possible changes in the data distribution.

\begin{algorithm}[h!]
	\caption{\small Adaptive Bandit with Context-Dependent Embeddings (ABaCoDE)}
	\label{alg:summary}
	\begin{algorithmic}[1]
    \STATE {\bfseries } \textbf{Input:} unlabeled dataset $\textbf{D}$, a set of unlabeled contexts for pre-training; $k$, the number of clusters (and corresponding embeddings); a Boolean variable $isOnline$.
    \STATE {\bfseries } \textbf{Initialization:} \STATE {\bfseries } \quad Cluster $\textbf{D}$ into $k$ clusters: \textbf{C} = $\{c_1, ..., c_k\}$
 \STATE {\bfseries} \quad For each cluster, train an autoencoder to construct a set of encoding functions (embeddings):    \textbf{E} = {$e_1, ..., e_k$} 
 \STATE {\bfseries} \quad Initialize the contextual Thompson Sampling parameters of bandit $B$ (line 1 in Alg. 1).
    \STATE {\bfseries } \textbf{while} there is a next data mini-batch $\textbf{M}$, \textbf{do}
    \STATE {\bfseries } \quad  \textbf{foreach} $x_t$ from  $\textbf{M}$ \textbf{do}
    \STATE {\bfseries }  \quad   \quad  \textbf{if} $isOnline$  \textbf{then}  $updateCluster(\textbf{C}, x_t, c_j)$ 
     \STATE {\bfseries } \quad   \quad  $e = selectEmbedding(c_j$)
   \STATE {\bfseries }   \quad   \quad $z=e(x_t)$ (encoded context/representation)
    \STATE {\bfseries }    \quad\quad  $contextualBandit(B,z)$
   (lines 4-7 in Alg. 1) \\
      \quad \textbf{end}
    \STATE {\bfseries }   \quad \textbf{if} $not (isOnline)$ \textbf{then} $recomputeClusters(\textbf{C}, \textbf{B}$)
    \STATE {\bfseries }   \quad$updateEmbedding(\textbf{M},\textbf{C}$) \\
\textbf{end}
    \end{algorithmic}
\end{algorithm}

The initialization step consists of clustering the pre-training dataset $\textbf{D}$ into $k$ clusters (line 3),  training an autoencoder for each cluster, which results into $k$ encoding (embedding) functions (line 4), and initializing 
parameters of the contextual Thompson Sampling bandit, used later to make classification decisions based on embedded context (line 5).

Next, the algorithm switches to the online mode, processing  an online stream of incoming samples (contexts). 
As mentioned above,  we assume that at the end of each fixed-length time window, i.e. a  fixed-size  mini-batch of data, we update our embeddings.

Within each data mini-batch $\textbf{M}$ (line 7),  once the next input sample ${\bf x}_i$ arrives, it is first assigned to one of the existing clusters $c_j$ (line 8), associated with the corresponding embedding function $e_j$.  Next, an online clustering is performed if $isOnline$ is true, i.e. the centroid of the cluster $c_j$ is recomputed, but no changes are made to other clusters (line 9). Otherwise, there are no changes to clusters, until the end of the  batch, as we will see shortly.  Based on the cluster assignment $c_j$, the corresponding  embedding function $e_j$ is used to compute the representation vector $z$ for given input ${\bf x}_i$ (line 10); given the context $z$, the contextual bandit $B$ makes a decision (line 11), obtains the reward $r_i$  (line 12), and updates its parameters (line 13) using the contextual Thompson Sampling described in the previous section.

After the end of the  mini-batch $\textbf{M}$ is reached (line 14), if $isOnline$ was false, the clusters will be recomputed from scratch using all data points received so far (however, no such re-clustering is performed if the online clustering was selected). Finally, the embeddings (i.e., their corresponding autoencoder parameters) are updated  respectively using the updated set of clusters $\textbf{C}$.

In the next section, we present empirical results comparing both online and offline clustering methods outlined above with two baseline approaches:
	\begin{itemize}
	\item \textit{Contextual  Bandit (CB)}: as the baseline, we use the standard contextual multi-armed bandit  with Thompson Sampling, based on the  raw input (i.e., no embeddings).
    	\item \textit{universal embedding (uE)}:  a universal embedding denotes a single embedding computed based on all data, and always recomputed to include the data from   the most recent mini-batch; no clustering is performed.
	\item \textit{mini-batch embedding (mE)}:   this is our offline clustering approach presented in Algorithm \ref{alg:summary}, when $isOnline$ is $false$. 
       \item \textit{online embedding (oE)}: this is the online version of our algorithm described above, i.e.  $isOnline$ is $true$. 
	\end{itemize}

\section{Empirical Evaluation}

\subsection{Datasets}

We evaluated our approach on four imaging datasets: MNIST \citep{lecun1998mnist}, STL-10 \citep{coates2011importance}, CIFAR-10 \citep{coates2011analysis}, Caltech-101 Silhouettes-28 \citep{griffin2007caltech} and Warfarin \citep{international2009estimation} (for details of each dataset, see Table \ref{table:datasets}). To simulate an online data stream, we draw samples from each dataset sequentially, starting from the beginning each time we draw the last sample. At each round, the algorithm receives  reward 1 if the instance is classified correctly, and 0 otherwise.
We compute the total number of classification errors as a performance metric.  However,  Warfarin dataset is different, as it was actually produced in a real bandit setting, rather than classification setting.

{\em Bandit vs. classification feedback: important distinction.} It is important to keep in mind that the bandit feedback (correct/incorrect classification) makes the classification problem significantly more challenging,  as compared to the standard supervised learning, since the true label is never revealed in bandit setting unless the classification is correct. Thus, the classification accuracy in a bandit setting is expected to be lower than in the supervised learning setting -- which is not due to  inferiority of   bandit decision making algorithm versus classifiers, but due to increased problem difficulty, i.s. the lack of feedback about what the correct decision should have been. Recall that such bandit feedback is often a much more realistic model of agent's interaction with the world, especially in online decision making applications such as online advertisement, clinical trials, and so on,
which do not fit into classification framework.

However, for empirical evaluation purposes, it is  common  to use available classification datasets to simulate an online environment with the bandit feedback  (i.e., simulating the situation where the bandit receives, for example, 1 or 0 for correct or incorrect decision, but is not told what the correct decision should have been when  he receives 0; such feedback is different from standard online classification feedback in case of non-binary classification)). We use several classification datasets here  for such simulations.
  
\begin{table}[t]
	\centering
	\caption{Datasets}
	\resizebox{0.99\columnwidth}{!}{
		\begin{tabular}{ l | c | c | r | r }
		  Datasets & History & Instances    & Features   & Classes \\ \hline
			MNIST & 10 000 & 20 000 & 784  &  10\\
			STL-10 & 20 000 & 10 000 & 1 000 & 10 \\
			CIFAR-10 & 2 000 & 10 000 & 3 072 & 10\\
			Caltech-101 S & 671 & 8 000 & 784 & 101\\ 
			Warfarin & 528 & 5 000 & 93 & 3\\ 
            \hline
            mix: MNIST/Warfarin & 10 528 & 10 000 & 93 & 13 \\
            \hline
        \end{tabular}
	}
	\label{table:datasets}
\end{table}

We now describe some details of the experiments. For MNIST, we took 10,000 samples from the original test dataset (clearly, not using them later for testing) to pre-train the encodings, and 60,000 samples from the training dataset to simulate the online bandit with 10 arms corresponding to different digits. For STL-10,  100,000 samples of unlabeled data are used to pre-train the encodings; then the 5,000 test samples together with  8,000 training samples are combined to simulate the online bandit, again with 10 different arms corresponding to image classes\footnote{To speed up the computation, we squeezed input 27648-dimensional vectors  into 1000-dimensional ones via linear stretching.}.  
For Caltech-101 Silhouettes-28 dataset, out of the  original 8671 samples, 671 are used for pre-training and 8000 for online learning with 101 different arms (class labels). For CIFAR-10 dataset,  10,000 test set samples are used for pre-training, and  50,000 training samples are left for the online bandit with 10 arms (classes). For Warfarin dataset,  528 test set samples are used for pre-training, and  5,000 training samples are left for the online bandit with 3 arms (classes).

\subsection{Nonstationary Environments}
We simulated  several types of nonstationarity using the above datasets (as in \citep{lin2020diabolic}). As mentioned before, we assume that the input data arrive in batches, and the data distribution (i.e., the joint distribution of the context and reward) may change across those batches, while remaining stationary within each batch.
We used the batch size of 1,000, and varied the number of embeddings $k$, using $k=2$, 4, or 8, presenting average results over all $k$.

\subsubsection{Nonstationary context: varying cluster distribution} To simulate changes in the context (input) distribution, we first   clustered all samples in the corresponding  pre-training data subset   into $k$ clusters. Next, we generate a sequence of batches, where each  batch   contained a certain fraction of samples from different clusters, and these fractions were changing across the batches, i.e. the probability distribution of cluster membership was changing, simulating nonstationary input.

\subsubsection{Nonstationary context: negative images}
Another type of input  nonstationarity involved introducing negative images as inputs with same semantics but different textures. Namely,  with probability $p$, the negative image of the original image was  presented as an input. 
Experiments were performed in two settings:  \textit{half} ($p = 0.5$) and \textit{rand} ($0 < p < 1$ randomly assigned for each mini-batch), in both stationary and nonstationary context conditions, with both shuffled and unshuffled rewards (described later).

\subsubsection{Nonstationary reward: multi-task environment} Another type of nonstationarity was  assuming that  input samples may come from different domains (tasks), and thus can be associated with different subsets of labels (arms). For example, we  combined 5,000 randomly selected training  samples from each of the two selected domains, MNIST and  Warfarin datasets, and extended the set of possible labels (arms) to include 10 labels from MNIST and 3 labels from Warfarin. We used  linear stretching to make the  input dimensions equal across the two domains. The algorithm had to assign a label to each input without any information about which domain the input came from.

\subsubsection{Nonstationary reward: shuffled class labels}
 We further explored the multi-task setting by introducing a different type of nonstationary reward, where 
 the class labels   were shuffled, i.e. randomly permuted, in each batch.





\subsection{Results}
\begin{figure*}[t]
\begin{multicols}{3}
    \includegraphics[width=1\linewidth]{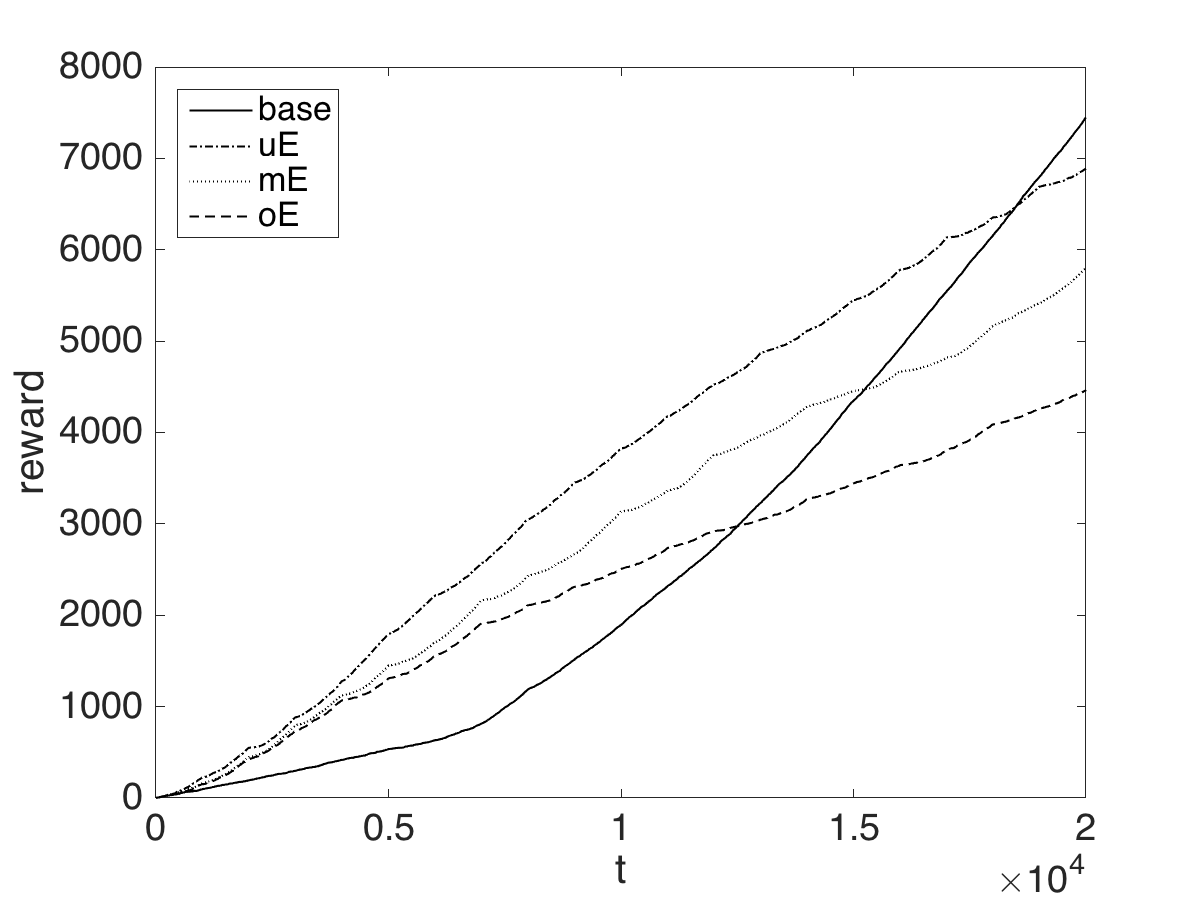}\par\caption{MNIST unshuffled, k = 2}\label{fig:MNIST-un}
    \includegraphics[width=1\linewidth]{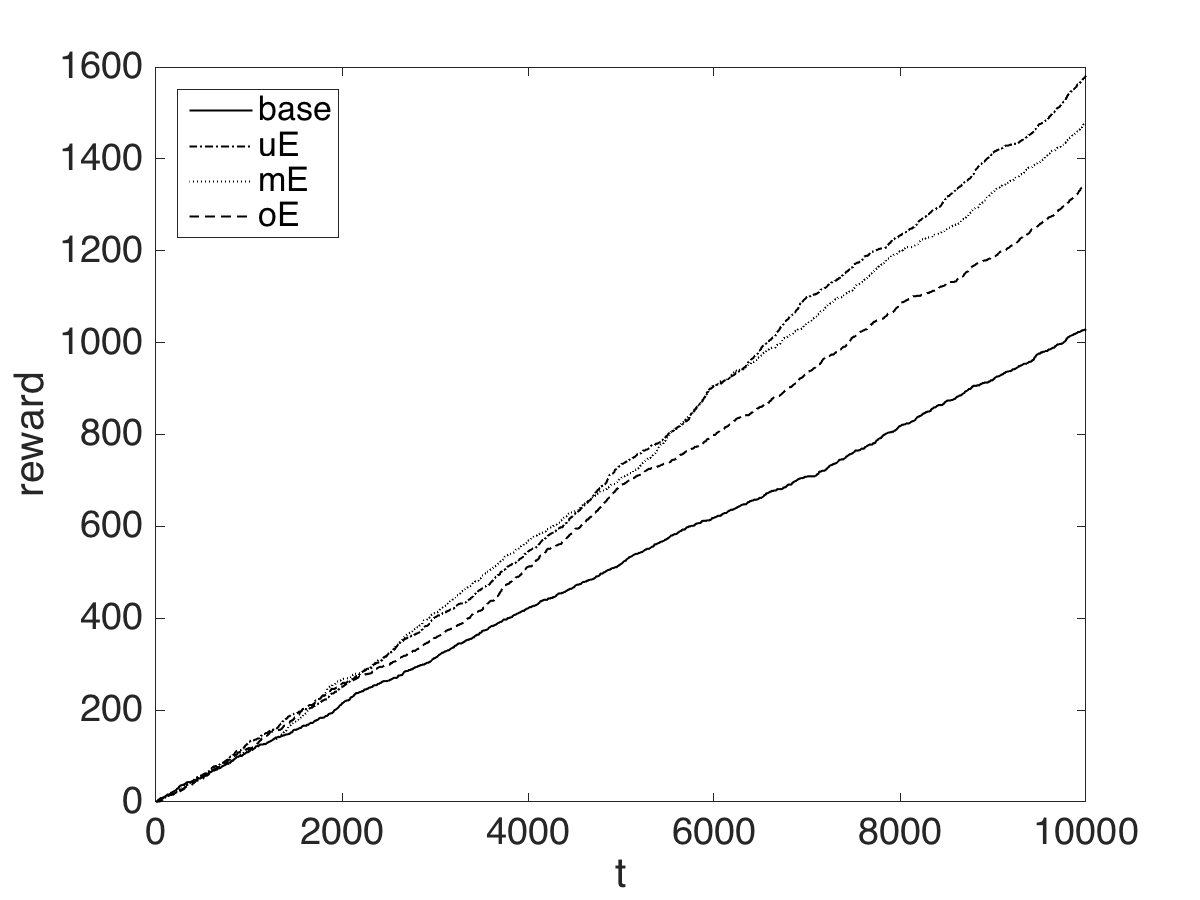}\par\caption{STL-10 unshuffled, k = 2}\label{fig:STL-un}
    \includegraphics[width=1\linewidth]{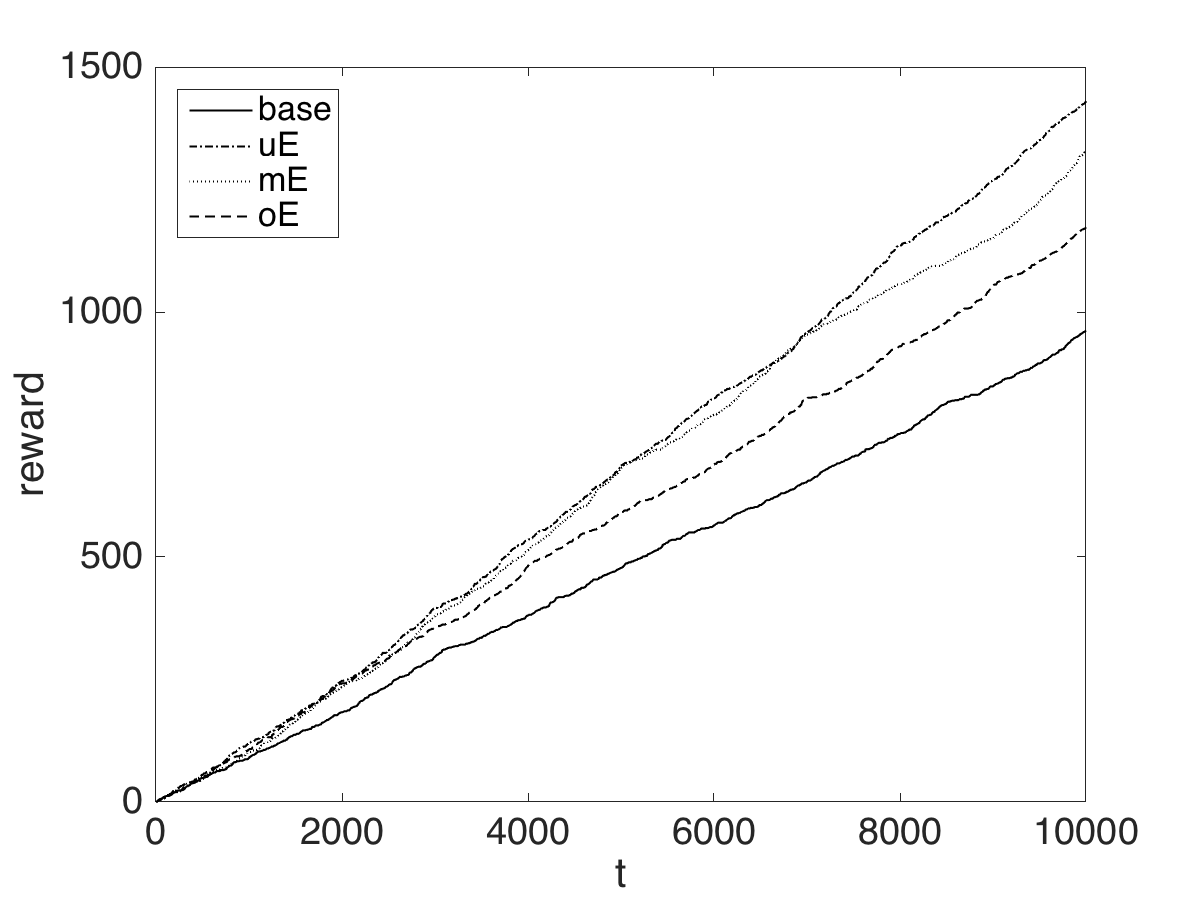}\par\caption{CIFAR-10 unshuffled, k = 2}\label{fig:CIFAR-un}
\end{multicols}
\end{figure*}

We explored different combination of the above nonstationarities.
Table \ref{table:nonstatAccUnshuffled} summarizes our results for the nonstationary context due to varying cluster distribution, and for mixed-domain (multi-task) settings, with {\em unshuffled reward function}. As we can see, on three out of six datasets, baseline was still outperforming our embeddings.
However, if we consider  the mean accuracy in the entire set of experiments, the top three algorithms were: \textit{universal embedding} (mean accuracy 28.83$\%$), \textit{baseline} (mean accuracy 27.78$\%$), \textit{mini-batch embedding} (mean accuracy 27.58$\%$),  respectively, suggesting the advantage of representation learning (embedding computation). Moreover, if we take a look at the whole iteration history, for example, for MNIST dataset (Figure \ref{fig:MNIST-un}), we observe that initially, the baseline CB (solid line) is considerably worse than embedding-based approaches, and requires a   large number of iteration to finally catch up with them. Figures \ref{fig:STL-un} and \ref{fig:CIFAR-un} show the history of reward accumulation for the STL-10 and CIFAR-10, demonstrating that the baseline is consistently outperformed by embedding selection methods.


\begin{table}[tbh]
	\centering
	\caption{Nonstationary Environment with   Unshuffled Labels}
	\resizebox{0.92\columnwidth}{!}{
		\begin{tabular}{ l | l | l | l  | l }
			Datasets & baseline & uE & mE &  oE  \\ \hline 
			MNIST & \textbf{37.24} & 34.44 & 29.00 & 22.32 \\
			STL-10 &  10.29 & \textbf{15.81} & 14.77 & 13.43 \\
			CIFAR-10 & 9.62 & \textbf{14.30} & 13.30 & 11.73 \\
			Caltech-101 S & \textbf{1.18} & 1.14 & 1.09 & 1.06 \\
            Warfarin & \textbf{62.58} & 56.70 & 56.10 & 56.92 \\
			\hline
            mix: MNIST/Warfarin & 45.76 & 50.58 & \textbf{51.21} & 47.74\\
            \hline
		\end{tabular}
	}  
\label{table:nonstatAccUnshuffled}
\end{table}

\begin{table}[tbh]
	\centering
	\caption{Nonstationary Environment with Shuffled Labels}
	\resizebox{0.92\columnwidth}{!}{
		\begin{tabular}{ l | l | l | l  | l }
			Datasets & baseline & uE & mE &  oE  \\ \hline 
			MNIST & 12.19 & \textbf{33.75} & 29.04 & 23.83 \\
			STL-10 &  10.05 & \textbf{16.64} & 15.10 & 12.77 \\
			CIFAR-10 & 10.23 & \textbf{14.83} & 13.13 & 11.60 \\
			Caltech-101 S & 1.00 & 1.09 & 1.23 & \textbf{1.30} \\
            Warfarin & 40.66 & \textbf{55.10} & 50.56 & 54.44 \\
			\hline
            mix: MNIST/Warfarin & 23.54 & 49.33 & \textbf{50.67} & 49.15 \\
            \hline
		\end{tabular}
	}  
\label{table:nonstatAccShuffled}
\end{table}

\begin{table}[tbh]
	\centering
	\caption{Negative Environment with Unshuffled Labels}
	\resizebox{0.92\columnwidth}{!}{
		\begin{tabular}{ l | l | l | l | l }
			Datasets & baseline & uE & mE &  oE  \\ \hline 
			MNIST half-stat & 13.50 & 14.70 & 14.02 & \textbf{16.18} \\
			MNIST rand-stat & 13.72 & 17.14 & 15.53 & \textbf{17.70}  \\
			MNIST half-nonStat & 14.45 & 25.09 & 23.82 & \textbf{26.90} \\
			MNIST rand-nonStat & 14.05 & 24.38 & 25.90 & \textbf{28.43}   \\  \hline
            STL-10 half-stat & 10.06 & \textbf{10.42} & 10.33 & 10.04 \\
            STL-10 rand-stat & 9.77 & \textbf{12.34} & 12.33 & 10.41 \\
            STL-10 half-nonStat & 9.88 & 10.99 & \textbf{12.29} & 11.56 \\
            STL-10 rand-nonStat & 9.85 & 12.99 & \textbf{13.67} & 11.55 \\
			\hline
            Caltech-101 S half-stat &  0.98 & \textbf{10.04} & 7.98 & 6.94 \\
            Caltech-101 S rand-stat & 0.94 & 10.93 & 8.40 & \textbf{11.68}  \\
            Caltech-101 S half-nonStat & 1.04 & 1.20 & \textbf{1.23} & 0.96 \\
            Caltech-101 S rand-nonStat & 0.96 & 1.09 & \textbf{1.20} & 0.99 \\
			\hline
		\end{tabular}
	}
	\label{table:AccUnshuffledNeg}
\end{table}

\begin{table}[tbh]
	\centering
	\caption{Negative Environment with Shuffled Labels}
	\resizebox{0.92\columnwidth}{!}{
		\begin{tabular}{ l | l | l | l | l }
			Datasets & baseline & uE & mE & oE  \\ \hline 
			MNIST half-stat & 10.22  & 14.59 & 13.86 & \textbf{14.79} \\
			MNIST rand-stat & 9.87 & \textbf{17.84} & 14.35 & 17.32  \\
			MNIST half-nonStat & 10.78 & 23.02 & 22.33 & \textbf{26.84} \\
			MNIST rand-nonStat & 11.27 & 27.34 & 24.87 & \textbf{28.36} \\ \hline
            STL-10 half-stat & 9.66 & \textbf{11.51} & 10.73 & 10.60 \\
            STL-10 rand-stat & 9.95 & 11.44 & \textbf{12.37} & 11.17 \\
            STL-10 half-nonStat & 10.31 & 11.86 & \textbf{13.17} & 11.19  \\
            STL-10 rand-nonStat & 9.2 & \textbf{12.62} & 12.49 & 11.59  \\
			\hline
            Caltech-101 S half-stat &  1.11 & \textbf{8.71} & 6.21 & 7.93 \\
            Caltech-101 S rand-stat & 0.94 & \textbf{10.36} & 9.11 & 3.38 \\
            Caltech-101 S half-nonStat & 1.06 & 1.03 & 1.00 & \textbf{1.29} \\
            Caltech-101 S rand-nonStat & 1.08 & 1.05 & 1.13 & \textbf{1.16} \\
			\hline
		\end{tabular}
	}
	\label{table:AccShuffledNeg}
\end{table}

\begin{figure*}[t]
\begin{multicols}{3}
    \includegraphics[width=1\linewidth]{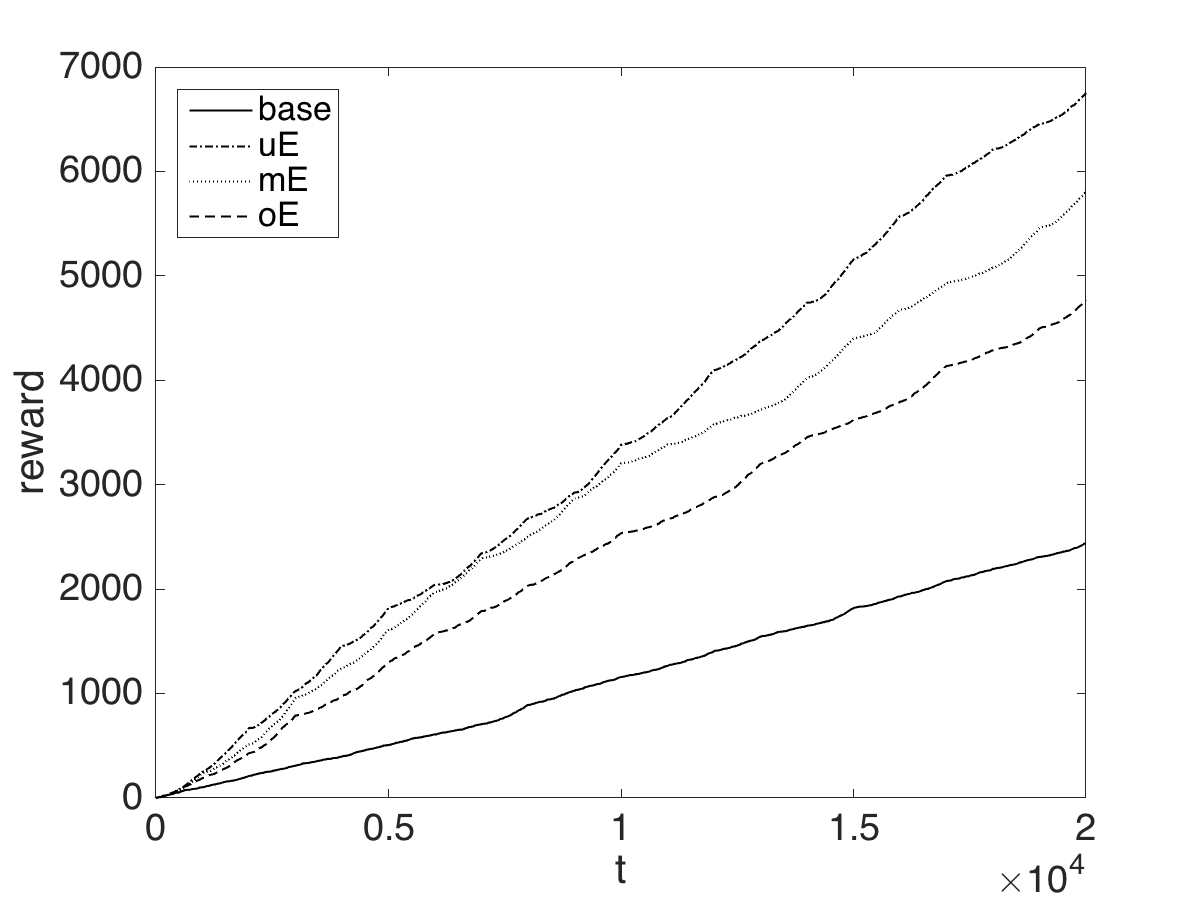}\par\caption{MNIST shuffled, k = 2 }\label{fig:MNIST-sh}
    \includegraphics[width=1\linewidth]{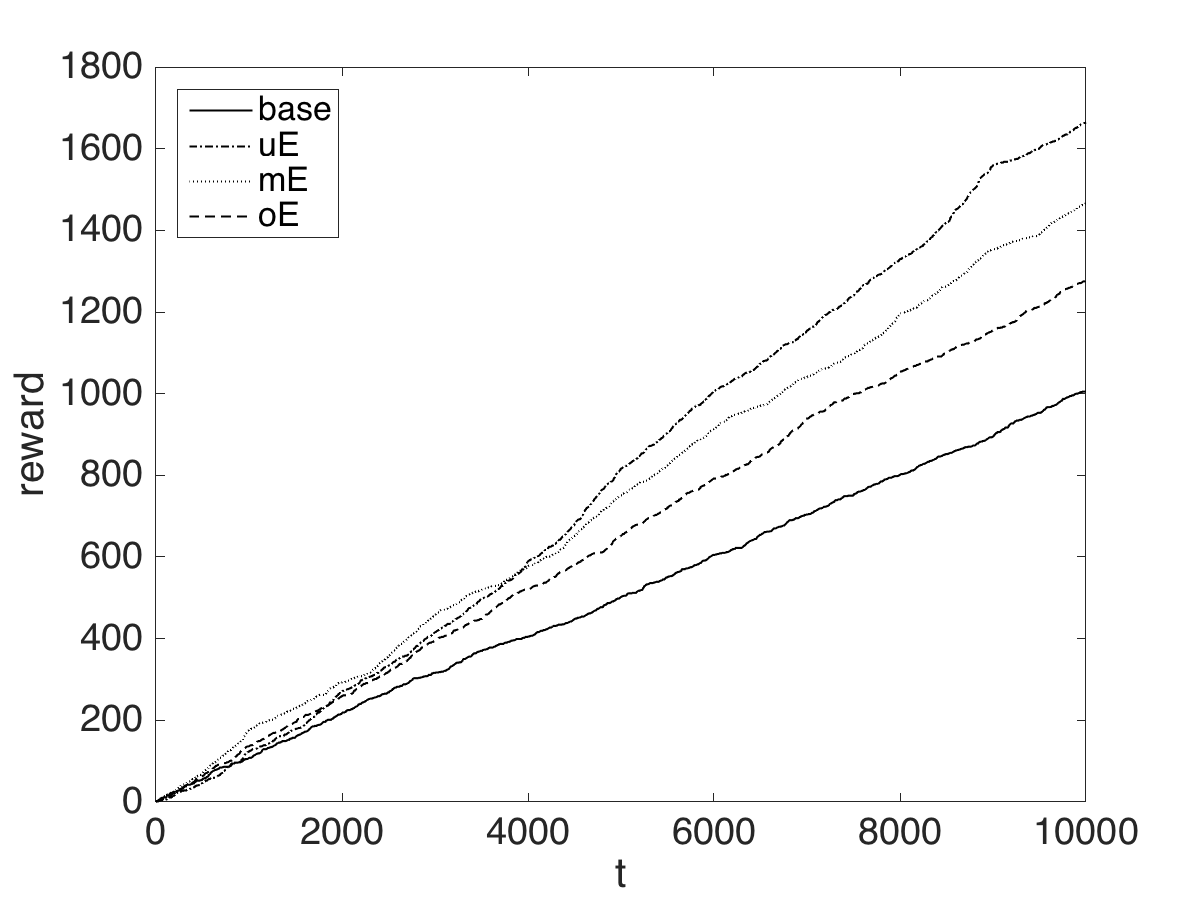}\par\caption{STL-10 shuffled, k = 2 }\label{fig:STL-sh}
    \includegraphics[width=1\linewidth]{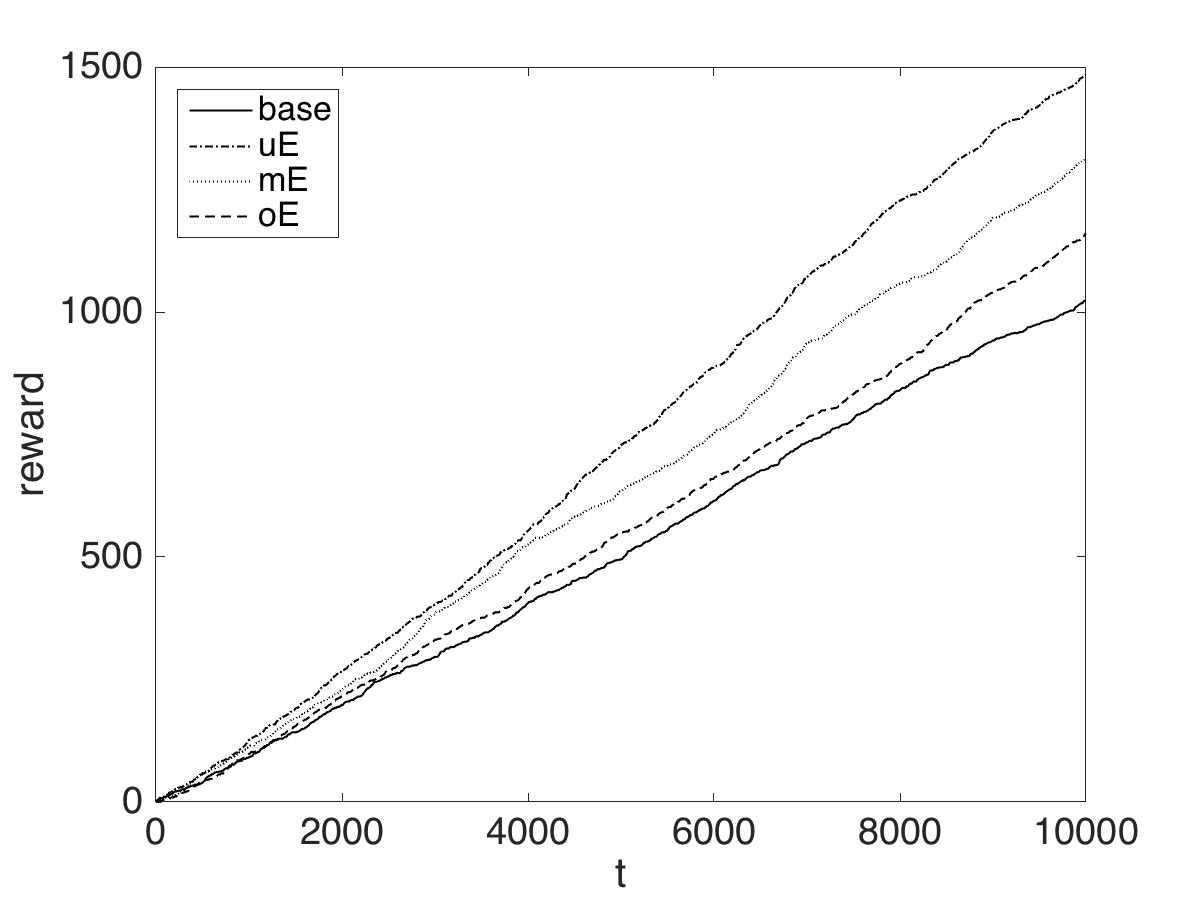}\par\caption{CIFAR-10 shuffled, k = 2} \label{fig:CIFAR-sh}
\end{multicols}
\end{figure*}



\begin{figure*}[t]
\begin{multicols}{4}
    \includegraphics[width=1\linewidth]{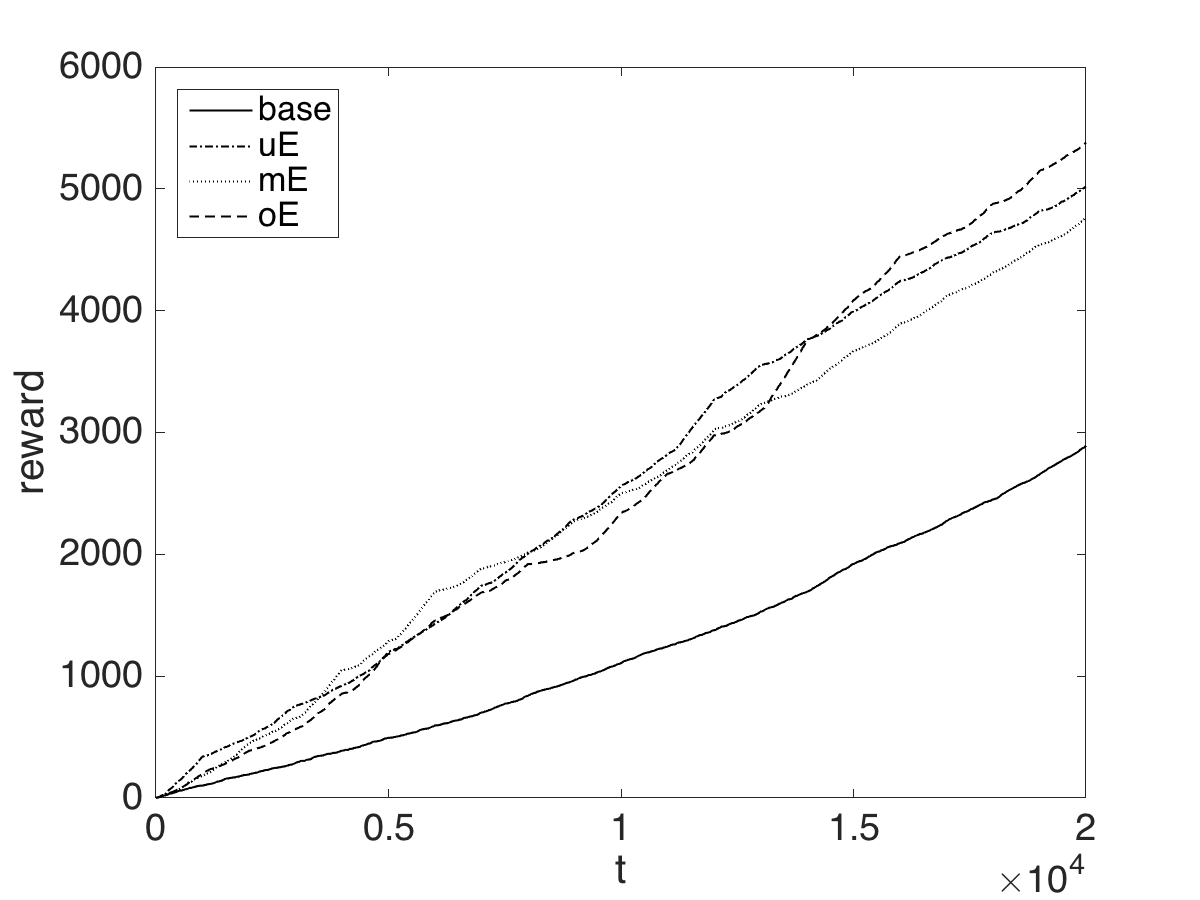}\par\caption{MNIST unshuffled half-nonStat}\label{fig:MNIST-un-half-nonStat}
    \includegraphics[width=1\linewidth]{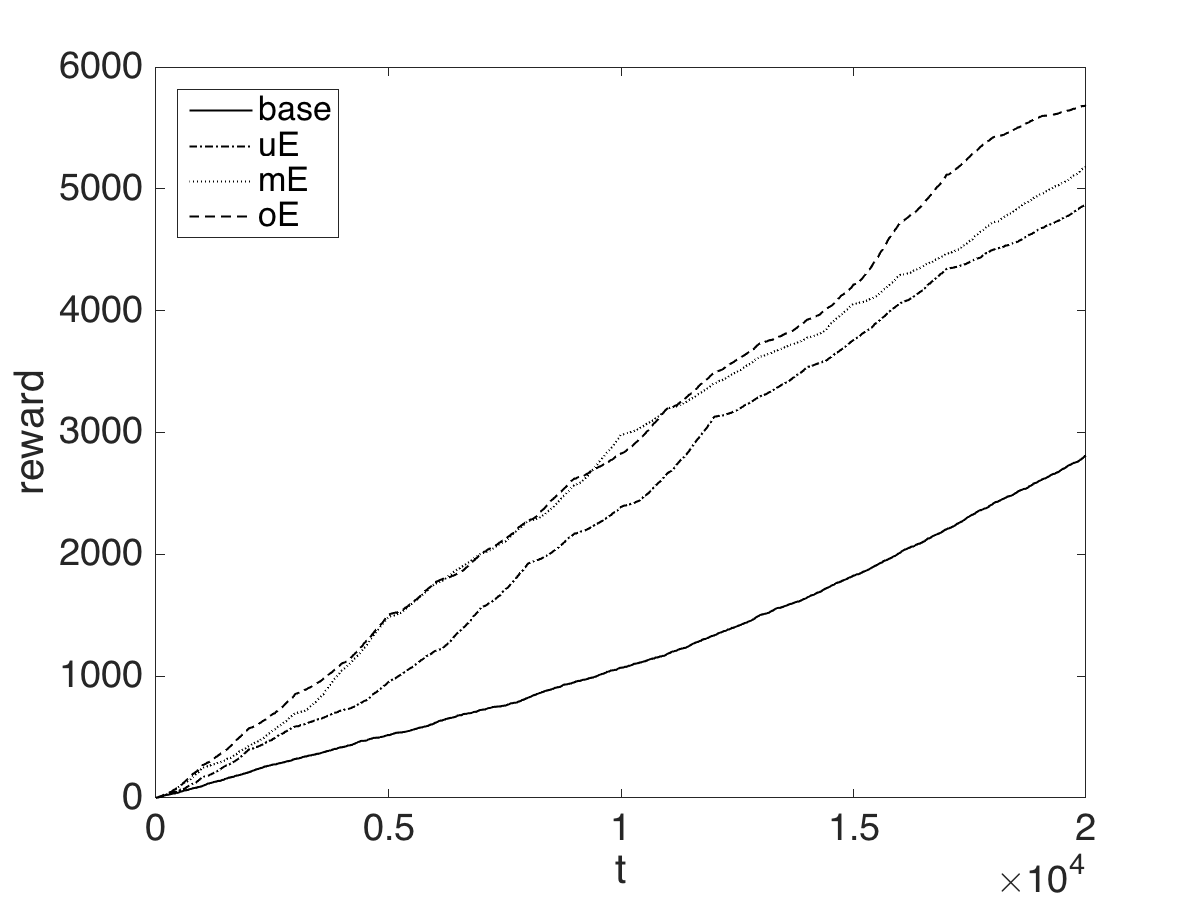}\par\caption{MNIST unshuffled rand-nonStat}\label{fig:MNIST-un-rand-nonStat}
    \includegraphics[width=1\linewidth]{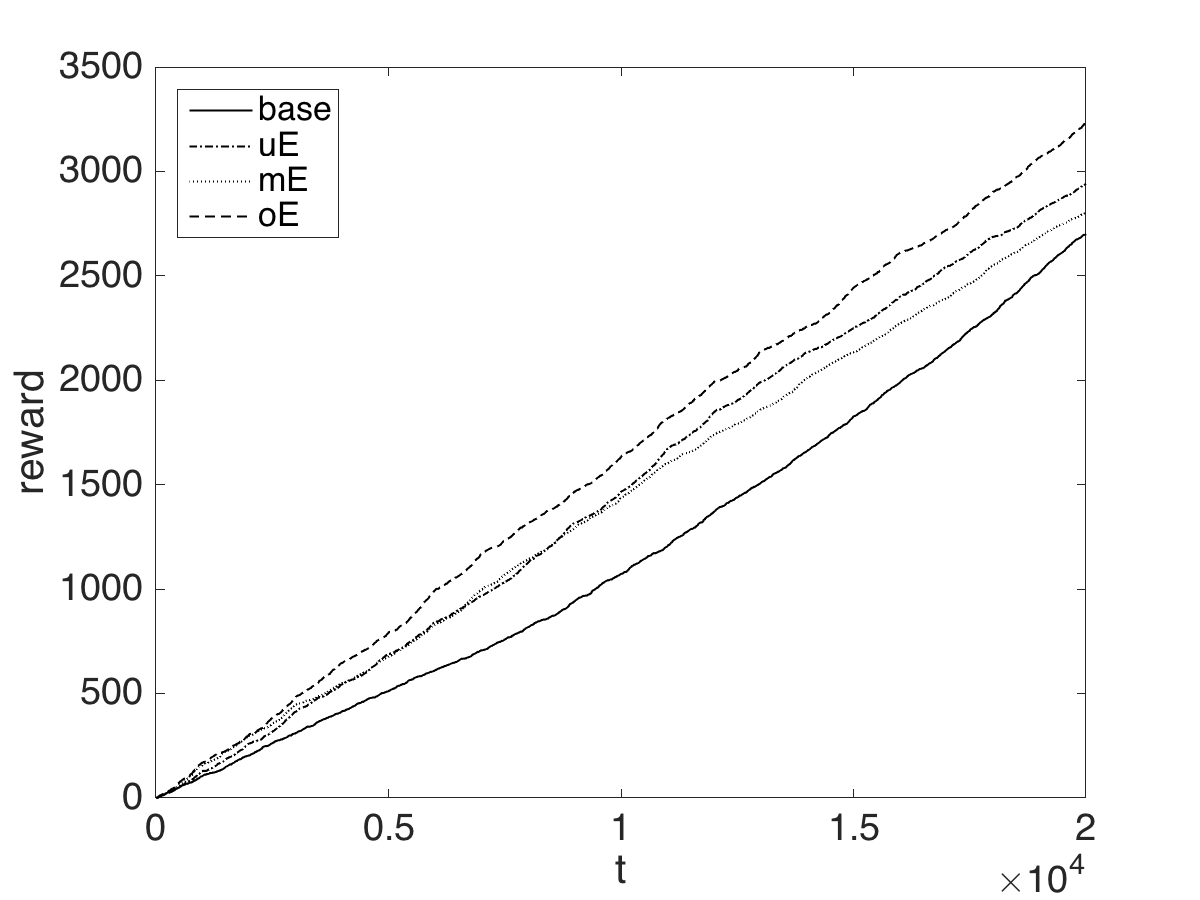}\par\caption{MNIST unshuffled half-stat}\label{fig:MNIST-un-half-stat}
        \includegraphics[width=1\linewidth]{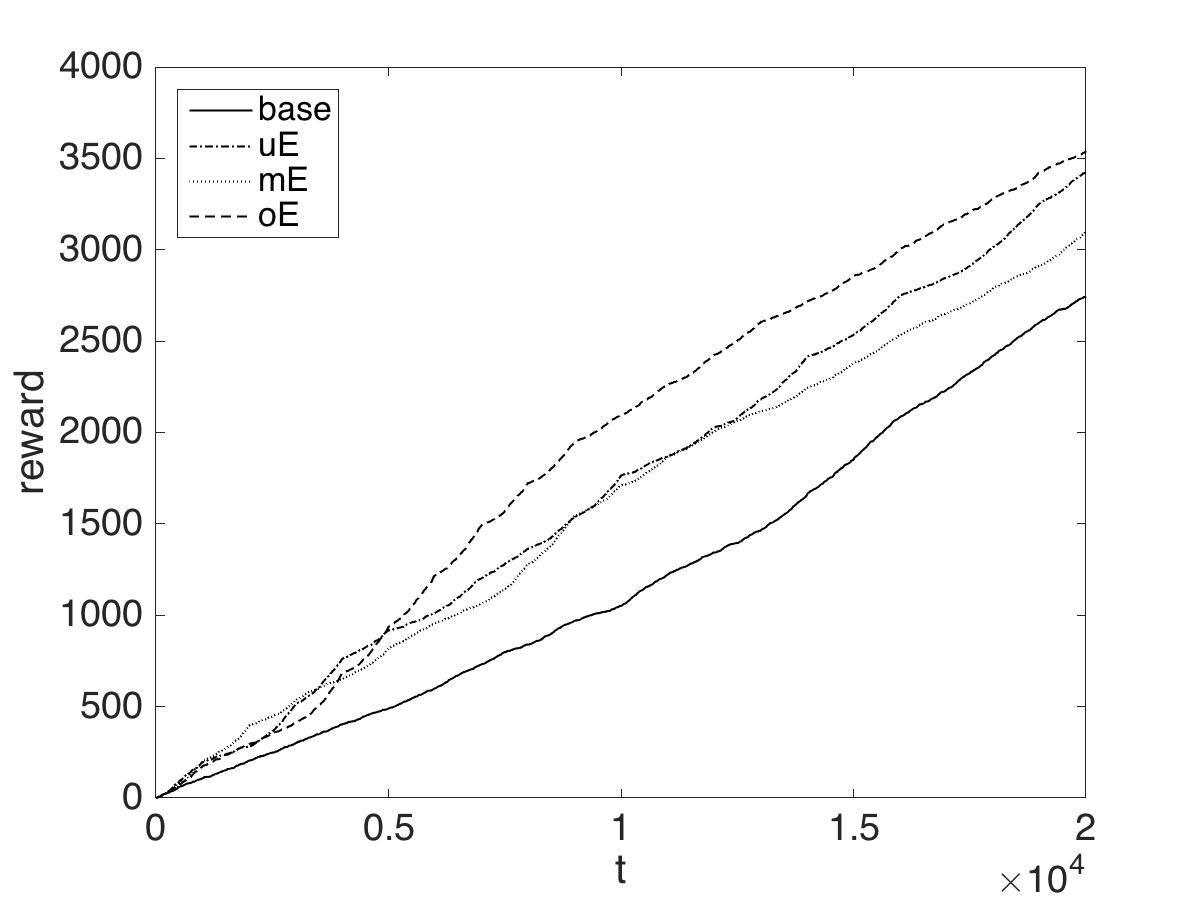}\par\caption{MNIST unshuffled rand-stat}\label{fig:MNIST-un-rand-stat}
\end{multicols}
\end{figure*}

\begin{figure*}[t]
\begin{multicols}{4}
    \includegraphics[width=1\linewidth]{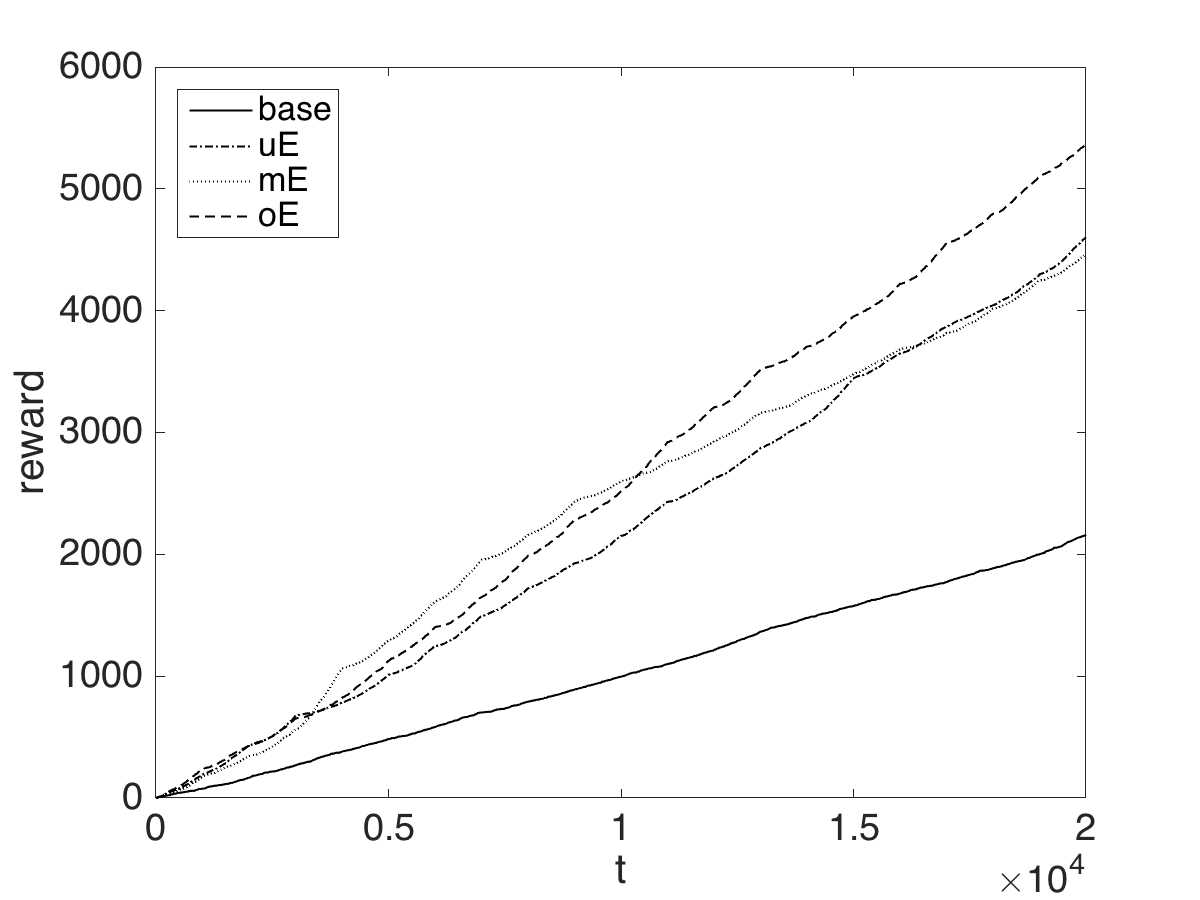}\par\caption{MNIST shuffled half-nonStat}\label{fig:MNIST-sh-half-nonStat}
    \includegraphics[width=1\linewidth]{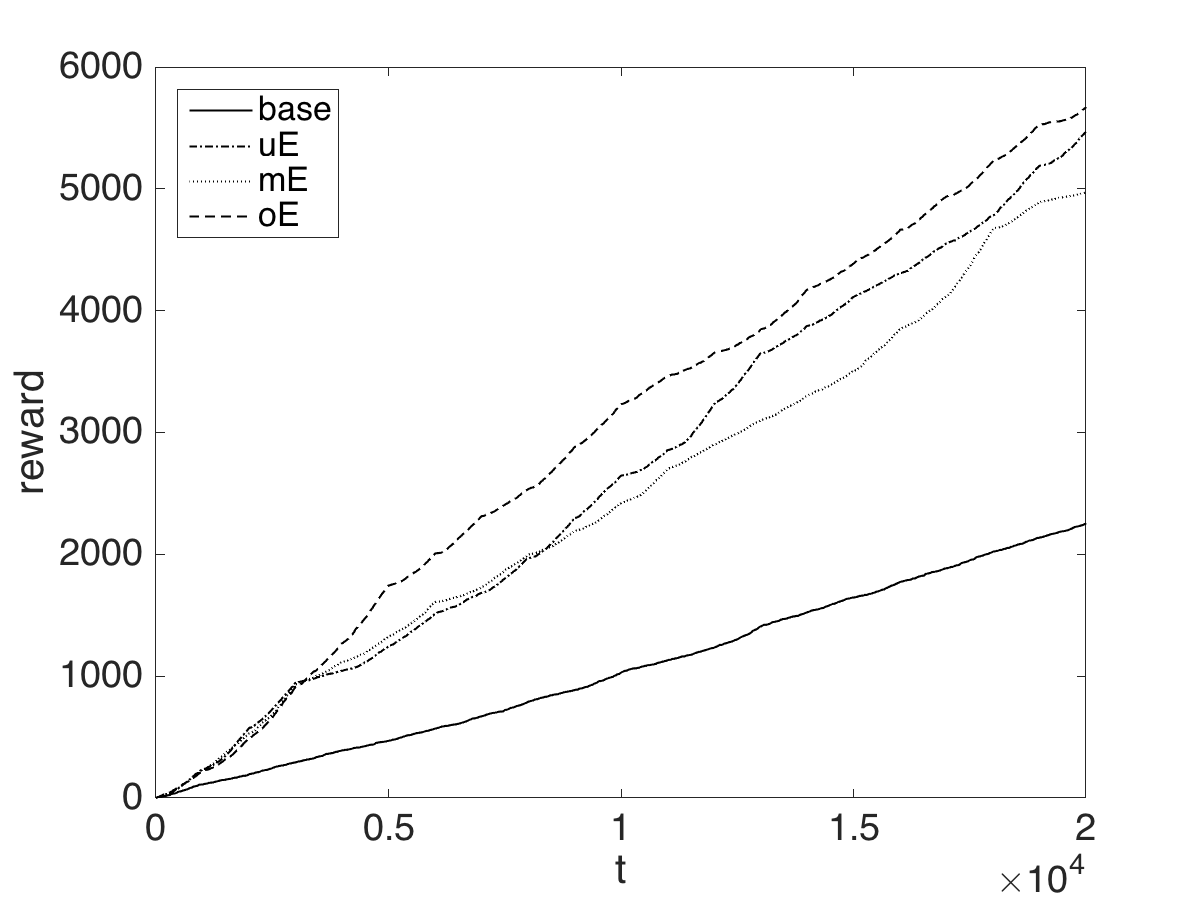}\par\caption{MNIST shuffled rand-nonStat}\label{fig:MNIST-sh-rand-nonStat}
    \includegraphics[width=1\linewidth]{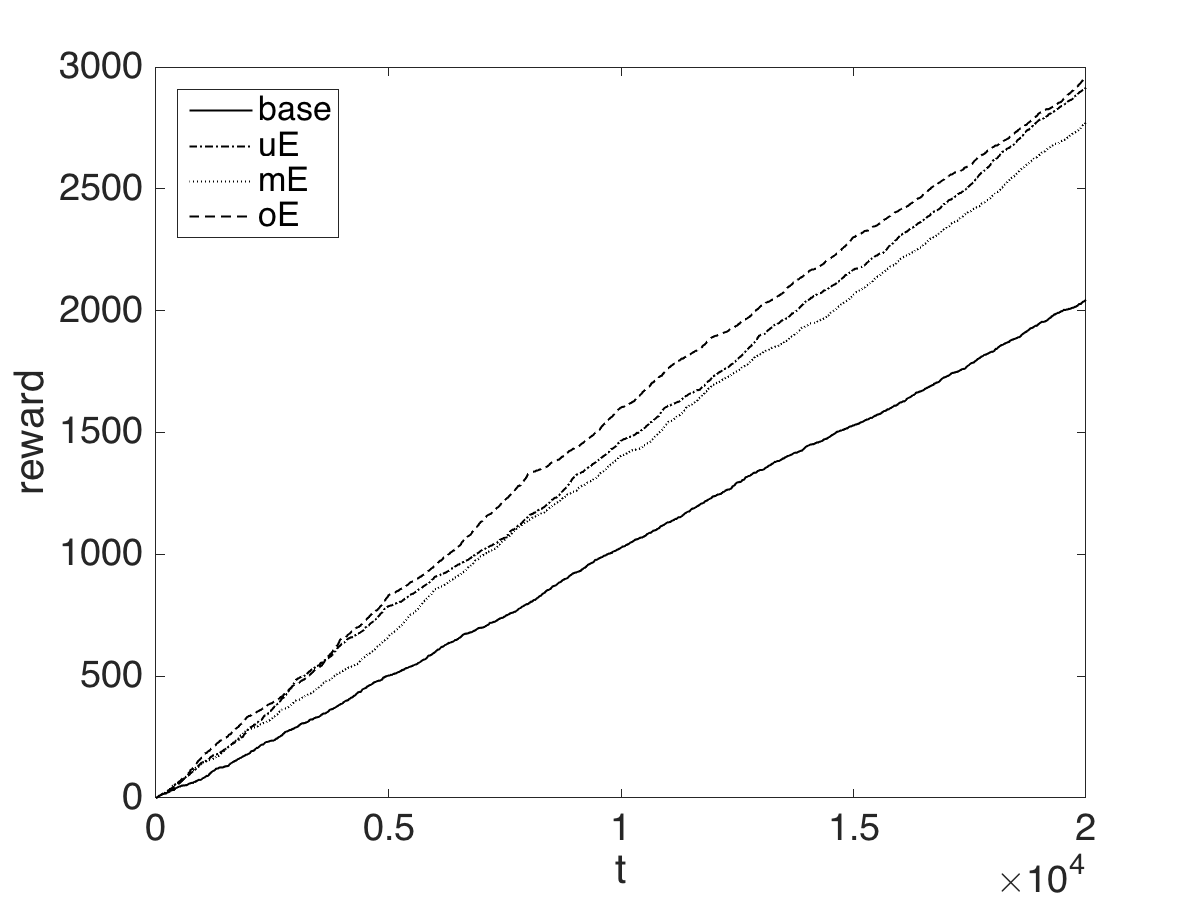}\par\caption{MNIST shuffled half-stat}\label{fig:MNIST-sh-half-stat}
        \includegraphics[width=1\linewidth]{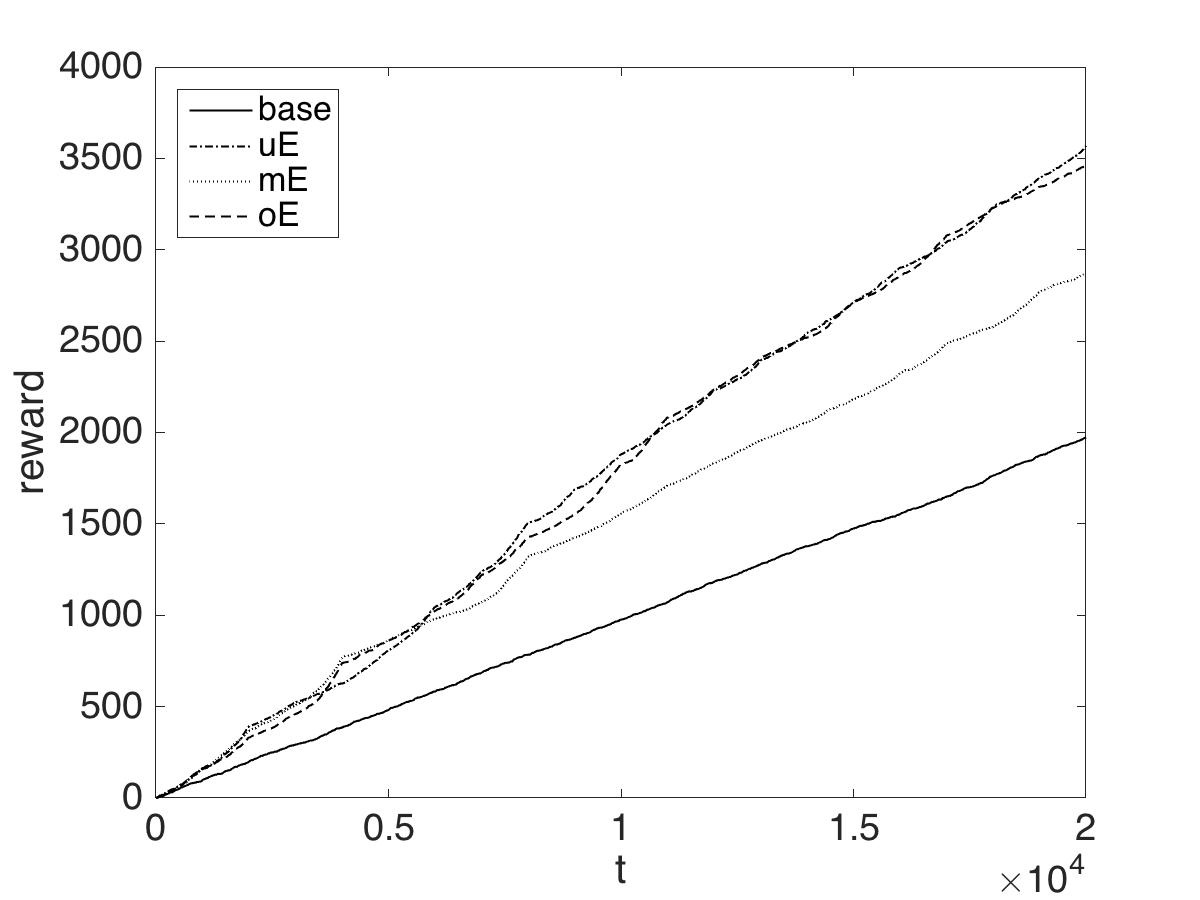}\par\caption{MNIST shuffled rand-stat}\label{fig:MNIST-sh-rand-stat}
\end{multicols}
\end{figure*}

Next, Table \ref{table:nonstatAccShuffled} summarizes our results  with {\em shuffled reward function}, for the nonstationary context due to varying cluster distribution, and for mixed-domain (multi-task) settings.
Based on the mean accuracy in the entire experiment, the top three algorithms were: \textit{universal embedding} (mean accuracy 28.46$\%$), \textit{mini-batch embedding} (mean accuracy 26.62$\%$), \textit{online embedding} (mean accuracy 25.52$\%$), respectively. Furthermore, in this experiment, our embedding-based approaches always outperformed the baseline, suggesting that in a setting where reward functions are nonstationary, in addition to the nonstationary input environment, the advantage of representation learning is quite significant, as compared to standard CB (mean accuracy 16.28$\%$). Note that, with nonstationary (shuffled) labels, the reward accumulated by the baseline CB remains significantly below the reward of   embedding-based approaches, at all iterations (Figures \ref{fig:MNIST-sh}-\ref{fig:CIFAR-sh}). Thus, in a more challenging setting with both context and reward nonstationarities, the embedding-based approaches clearly outperform the standard contextual bandit.

Table \ref{table:AccUnshuffledNeg} summarizes our results for the nonstationary online learning setting with {\em negative environments} and {\em unshuffled reward}. Based on the mean accuracy in the entire experiment, the top three algorithms were: \textit{online embedding} (mean accuracy 12.78$\%$), \textit{universal embedding} (mean accuracy 12.61$\%$), \textit{mini-batch embedding} (mean accuracy 12.23$\%$), respectively. Again, the embedding-based approaches are always superior to the baseline CB;   \textit{online embedding} achieved the best performance among all methods on MNIST, while universal and batch embeddings were taking their turns  outperforming the baseline on other datasets and settings.

Finally, Table \ref{table:AccShuffledNeg} summarizes our results for the nonstationary online learning setting with negative environments and shuffled reward function. Based on the mean accuracy in the entire experiment, the top three algorithms were: \textit{universal embedding} (mean accuracy 12.61$\%$), \textit{online embedding} (mean accuracy 12.14$\%$), \textit{mini-batch embedding} (mean accuracy 11.80$\%$), respectively, further confirming the advantage of adaptive encoding over standard CB (mean accuracy 7.12$\%$). In addition, the difference of textures under the same semantics introduced in this experiments demonstrated that embedding selection outperforms single universal embedding in most nonstationary cases. 

Figures \ref{fig:MNIST-un-half-nonStat}-\ref{fig:MNIST-sh-rand-stat} visualize the details of reward accumulation over time by different methods, on MNIST data and all the settings from the Tables \ref{table:AccUnshuffledNeg} and \ref{table:AccShuffledNeg}. The performance gap between the embedding-based approaches and the baseline is especially large in those settings. Furthermore, we can see that both   adaptive, context-dependent  embedding approaches  (oE and mE)  consistently ourperform the single-embedding approach (uE), with the {\em online embedding} emerging as the best one, especially with increasing number of iterations.

\section{Conclusions}
We  introduced an extension of the contextual bandit problem motivated by several real-world applications in non-stationary environments, including recommendation systems, health monitoring and medical diagnosis, and others. In this setting, which we refer to as Contextual Bandit with Representation learning and unlabeled History (CBRH), a set of unlabeled contexts is available prior to online decision making, which allows, instead of using the raw context, to learn context representations. Next, during the online phase,  embeddings are  selected adaptively, depending on each context, and   updated based on the contexts observed so far. We propose two specific algorithms for the CBRH problem, based on online and offline clustering, which combine online embedding selection and learning with contextual Thompson Sampling bandit. The algorithms are evaluated in several types of nonstationary environments and compared to the standard  contextual bandit, as well as universal (single) embedding, on several datasets. Overall, we  observe clear  advantages of the embedding-based approaches over the  standard contextual bandit;
moreover, the proposed adaptive embedding selection and learning methods frequently outperform the universal embedding in multiple nonstationary settings.

\bibliographystyle{named}  
\bibliography{references}

\begin{thebibliography}{}

\bibitem[\protect\citeauthoryear{Agrawal and Goyal}{2012}]{AgrawalG12}
Shipra Agrawal and Navin Goyal.
\newblock Analysis of thompson sampling for the multi-armed bandit problem.
\newblock In {\em {COLT} 2012 - The 25th Annual Conference on Learning Theory,
  June 25-27, 2012, Edinburgh, Scotland}, pages 39.1--39.26, 2012.

\bibitem[\protect\citeauthoryear{Agrawal and Goyal}{2013}]{AgrawalG13}
Shipra Agrawal and Navin Goyal.
\newblock Thompson sampling for contextual bandits with linear payoffs.
\newblock In {\em ICML (3)}, pages 127--135, 2013.

\bibitem[\protect\citeauthoryear{Allesiardo \bgroup \em et al.\egroup
  }{2014}]{AllesiardoFB14}
Robin Allesiardo, Rapha{\"{e}}l F{\'{e}}raud, and Djallel Bouneffouf.
\newblock A neural networks committee for the contextual bandit problem.
\newblock In {\em Neural Information Processing - 21st International
  Conference, {ICONIP} 2014, Kuching, Malaysia, November 3-6, 2014.
  Proceedings, Part {I}}, pages 374--381, 2014.

\bibitem[\protect\citeauthoryear{Auer and Cesa-Bianchi}{1998}]{AuerC98}
Peter Auer and Nicol{\`o} Cesa-Bianchi.
\newblock On-line learning with malicious noise and the closure algorithm.
\newblock {\em Ann. Math. Artif. Intell.}, 23(1-2):83--99, 1998.

\bibitem[\protect\citeauthoryear{Auer \bgroup \em et al.\egroup }{2002a}]{UCB}
Peter Auer, Nicol{\`o} Cesa-Bianchi, and Paul Fischer.
\newblock Finite-time analysis of the multiarmed bandit problem.
\newblock {\em Machine Learning}, 47(2-3):235--256, 2002.

\bibitem[\protect\citeauthoryear{Auer \bgroup \em et al.\egroup
  }{2002b}]{AuerCFS02}
Peter Auer, Nicol{\`o} Cesa-Bianchi, Yoav Freund, and Robert~E. Schapire.
\newblock The nonstochastic multiarmed bandit problem.
\newblock {\em SIAM J. Comput.}, 32(1):48--77, 2002.

\bibitem[\protect\citeauthoryear{Bart{\'o}k \bgroup \em et al.\egroup
  }{2014}]{bartok2014partial}
G{\'a}bor Bart{\'o}k, Dean~P Foster, D{\'a}vid P{\'a}l, Alexander Rakhlin, and
  Csaba Szepesv{\'a}ri.
\newblock Partial monitoring—classification, regret bounds, and algorithms.
\newblock {\em Mathematics of Operations Research}, 39(4):967--997, 2014.

\bibitem[\protect\citeauthoryear{Bouneffouf and
  F{\'{e}}raud}{2016}]{BouneffoufF16}
Djallel Bouneffouf and Rapha{\"{e}}l F{\'{e}}raud.
\newblock Multi-armed bandit problem with known trend.
\newblock {\em Neurocomputing}, 205:16--21, 2016.

\bibitem[\protect\citeauthoryear{Bouneffouf \bgroup \em et al.\egroup
  }{2017a}]{bouneffouf2017bandit}
Djallel Bouneffouf, Irina Rish, and Guillermo~A Cecchi.
\newblock Bandit models of human behavior: Reward processing in mental
  disorders.
\newblock In {\em International Conference on Artificial General Intelligence},
  pages 237--248. Springer, 2017.

\bibitem[\protect\citeauthoryear{Bouneffouf \bgroup \em et al.\egroup
  }{2017b}]{bouneffouf2017context}
Djallel Bouneffouf, Irina Rish, Guillermo~A Cecchi, and Rapha{\"e}l F{\'e}raud.
\newblock Context attentive bandits: Contextual bandit with restricted context.
\newblock In {\em Proceedings of IJCAI-2017}, 2017.

\bibitem[\protect\citeauthoryear{Chu \bgroup \em et al.\egroup
  }{2011}]{ChuLRS11}
Wei Chu, Lihong Li, Lev Reyzin, and Robert~E. Schapire.
\newblock Contextual bandits with linear payoff functions.
\newblock In Geoffrey~J. Gordon, David~B. Dunson, and Miroslav Dudik, editors,
  {\em AISTATS}, volume~15 of {\em JMLR Proceedings}, pages 208--214. JMLR.org,
  2011.

\bibitem[\protect\citeauthoryear{Coates and Ng}{2011}]{coates2011importance}
Adam Coates and Andrew~Y Ng.
\newblock The importance of encoding versus training with sparse coding and
  vector quantization.
\newblock In {\em Proceedings of the 28th International Conference on Machine
  Learning (ICML-11)}, pages 921--928, 2011.

\bibitem[\protect\citeauthoryear{Coates \bgroup \em et al.\egroup
  }{2011}]{coates2011analysis}
Adam Coates, Andrew Ng, and Honglak Lee.
\newblock An analysis of single-layer networks in unsupervised feature
  learning.
\newblock In {\em Proceedings of the fourteenth international conference on
  artificial intelligence and statistics}, pages 215--223, 2011.

\bibitem[\protect\citeauthoryear{Consortium and
  others}{2009}]{international2009estimation}
International Warfarin~Pharmacogenetics Consortium et~al.
\newblock Estimation of the warfarin dose with clinical and pharmacogenetic
  data.
\newblock {\em N Engl J Med}, 2009(360):753--764, 2009.

\bibitem[\protect\citeauthoryear{Gajane \bgroup \em et al.\egroup
  }{2016}]{gajanecorrupt}
Pratik Gajane, Tanguy Urvoy, and Emilie Kaufmann.
\newblock Corrupt bandits.
\newblock {\em EWRL}, 2016.

\bibitem[\protect\citeauthoryear{Griffin \bgroup \em et al.\egroup
  }{2007}]{griffin2007caltech}
Gregory Griffin, Alex Holub, and Pietro Perona.
\newblock Caltech-256 object category dataset.
\newblock 2007.

\bibitem[\protect\citeauthoryear{Lai and Robbins}{1985}]{LR85}
T.~L. Lai and Herbert Robbins.
\newblock Asymptotically efficient adaptive allocation rules.
\newblock {\em Advances in Applied Mathematics}, 6(1):4--22, 1985.

\bibitem[\protect\citeauthoryear{Langford and Zhang}{2008}]{langford2008epoch}
John Langford and Tong Zhang.
\newblock The epoch-greedy algorithm for multi-armed bandits with side
  information.
\newblock In {\em Advances in neural information processing systems}, pages
  817--824, 2008.

\bibitem[\protect\citeauthoryear{LeCun}{1998}]{lecun1998mnist}
Yann LeCun.
\newblock The mnist database of handwritten digits.
\newblock {\em http://yann. lecun. com/exdb/mnist/}, 1998.

\bibitem[\protect\citeauthoryear{Li \bgroup \em et al.\egroup }{2010}]{Li2010}
Lihong Li, Wei Chu, John Langford, and Robert~E Schapire.
\newblock A contextual-bandit approach to personalized news article
  recommendation.
\newblock In {\em {Proceedings of the 19th International Conference on World
  Wide Web (WWW2010)}}, pages 661--670. ACM, 2010.

\bibitem[\protect\citeauthoryear{Lin and Zhang}{2020a}]{lin2020speaker}
Baihan Lin and Xinxin Zhang.
\newblock Speaker diarization as a fully online learning problem in minivox.
\newblock {\em arXiv preprint arXiv:2006.04376}, 2020.

\bibitem[\protect\citeauthoryear{Lin and Zhang}{2020b}]{lin2020voiceid}
Baihan Lin and Xinxin Zhang.
\newblock {VoiceID} on the fly: A speaker recognition system that learns from
  scratch.
\newblock In {\em INTERSPEECH}, 2020.

\bibitem[\protect\citeauthoryear{Lin \bgroup \em et al.\egroup
  }{2019}]{lin2019split}
Baihan Lin, Djallel Bouneffouf, and Guillermo Cecchi.
\newblock Split {Q} learning: Reinforcement learning with two-stream rewards.
\newblock In {\em Proceedings of the Twenty-Eighth International Joint
  Conference on Artificial Intelligence, {IJCAI-19}}, pages 6448--6449.
  International Joint Conferences on Artificial Intelligence Organization, 7
  2019.

\bibitem[\protect\citeauthoryear{Lin \bgroup \em et al.\egroup
  }{2020a}]{lin2020online}
Baihan Lin, Djallel Bouneffouf, and Guillermo Cecchi.
\newblock Online learning in iterated prisoner's dilemma to mimic human
  behavior.
\newblock {\em arXiv preprint arXiv:2006.06580}, 2020.

\bibitem[\protect\citeauthoryear{Lin \bgroup \em et al.\egroup
  }{2020b}]{lin2020astory}
Baihan Lin, Guillermo Cecchi, Djallel Bouneffouf, Jenna Reinen, and Irina Rish.
\newblock A story of two streams: Reinforcement learning models from human
  behavior and neuropsychiatry.
\newblock In {\em Proceedings of the Nineteenth International Conference on
  Autonomous Agents and Multi-Agent Systems, {AAMAS-20}}, pages 744--752.
  International Foundation for Autonomous Agents and Multiagent Systems, 5
  2020.

\bibitem[\protect\citeauthoryear{Lin \bgroup \em et al.\egroup
  }{2020c}]{lin2020unified}
Baihan Lin, Guillermo Cecchi, Djallel Bouneffouf, Jenna Reinen, and Irina Rish.
\newblock Unified models of human behavioral agents in bandits, contextual
  bandits and rl.
\newblock {\em arXiv preprint arXiv:2005.04544}, 2020.

\bibitem[\protect\citeauthoryear{Lin}{2020a}]{lin2020diabolic}
Baihan Lin.
\newblock Diabolical games: Reinforcement learning environments for lifelong
  learning.
\newblock {\em under review}, 2020.

\bibitem[\protect\citeauthoryear{Lin}{2020b}]{lin2020semi}
Baihan Lin.
\newblock Online semi-supervised learning in contextual bandits with episodic
  reward.
\newblock In {\em Australasian Joint Conference on Artificial Intelligence}.
  Springer, 2020.

\bibitem[\protect\citeauthoryear{Mary \bgroup \em et al.\egroup
  }{2015}]{MaryGP15}
J{\'{e}}r{\'{e}}mie Mary, Romaric Gaudel, and Philippe Preux.
\newblock Bandits and recommender systems.
\newblock In {\em Machine Learning, Optimization, and Big Data - First
  International Workshop, {MOD} 2015}, pages 325--336, 2015.

\bibitem[\protect\citeauthoryear{Ororbia \bgroup \em et al.\egroup
  }{2015}]{ororbia2015online}
II~Ororbia, G~Alexander, C~Lee Giles, and David Reitter.
\newblock Online semi-supervised learning with deep hybrid boltzmann machines
  and denoising autoencoders.
\newblock {\em arXiv preprint arXiv:1511.06964}, 2015.

\bibitem[\protect\citeauthoryear{Thompson}{1933}]{T33}
W.R. Thompson.
\newblock On the likelihood that one unknown probability exceeds another in
  view of the evidence of two samples.
\newblock {\em Biometrika}, 25:285--294, 1933.

\bibitem[\protect\citeauthoryear{Villar \bgroup \em et al.\egroup
  }{2015}]{villar2015multi}
Sof{\'\i}a~S Villar, Jack Bowden, and James Wason.
\newblock Multi-armed bandit models for the optimal design of clinical trials:
  benefits and challenges.
\newblock {\em Statistical science: a review journal of the Institute of
  Mathematical Statistics}, 30(2):199, 2015.

\bibitem[\protect\citeauthoryear{Yver}{2009}]{Yver2009}
B.~Yver.
\newblock Online semi-supervised learning: Application to dynamic learning from
  radar data.
\newblock In {\em 2009 International Radar Conference "Surveillance for a Safer
  World" (RADAR 2009)}, pages 1--6, Oct 2009.

\end{thebibliography}

\end{document}